\documentclass[twoside,leqno,twocolumn]{article}

\usepackage[letterpaper]{geometry}
\usepackage{ltexpprt}
\usepackage{multirow}
\usepackage{graphicx}
\usepackage{amsmath}
\usepackage{booktabs}
\usepackage{amssymb}
\usepackage{svg}
\usepackage{array}
\usepackage{hyperref}
\usepackage{enumitem}  
\usepackage{chngcntr}
\counterwithin{table}{section}
\begin{document}

\title{\Large Deep Context-Aware Novelty Detection \thanks{ This work has been supported by a research grant by Science Foundation Ireland under grant number SFI/15/CDA/3520.}}
\author{Ellen Rushe \and Brian Mac Namee
\thanks{Insight Centre for Data Analytics, University College Dublin, Belfield, Dublin 4, Ireland }}

\date{}

\maketitle







\begin{abstract} \small\baselineskip=9pt
A common assumption of novelty detection is that the distribution of both ``normal" and ``novel" data are static. This, however, is often not the case---for example scenarios where data evolves over time or scenarios in which the definition of normal and novel depends on contextual information, both leading to changes in these distributions. This can lead to significant difficulties when attempting to train a model on datasets where the distribution of normal data in one scenario is similar to that of novel data in another scenario. In this paper we propose a context-aware approach to novelty detection for deep autoencoders to address these difficulties. We create a semi-supervised network architecture that utilises auxiliary labels to reveal contextual information and allow the model to adapt to a variety of contexts in which the definitions of normal and novel change.  We evaluate our approach on both image data and real world audio data displaying these characteristics and show that the performance of individually trained models can be achieved in a single model.

\end{abstract}

\section{Introduction}
\label{background}
Novelty detection is often framed as a task where the definition of ``normal", or data that has been ``seen" before, is static. Furthermore, the nature of what is ``novel" is often also assumed to be fixed. In real world scenarios, however, these assumptions are often invalid: the nature of normality and novelty may evolve as a function of time or depend on the \emph{context} in which data is observed \cite{chandola2009anomaly}. This paper addresses the latter scenario. There are many situations where an event that is normal in one context, may be considered novel in another. The data therefore exhibits two types of features, those that depend on different dynamic factors and those that are ubiqitous across all contexts, termed \textit{contextual attributes} and \textit{behavioral attributes} respectively \cite{chandola2009anomaly}.
One intuitive example of this type of anomaly might be the sound of an ice-cream truck on a sunny day versus the same sound in the middle of the night. A more concrete example arises in seizure detection using EEG, where cross-patient models are significantly more challenging to create than patient specific models \cite{thodoroff2016learning}. In this scenario, the patient to whom the data belongs defines the context within which a model should operate. This is easily obtained metadata that can be utilised in order to provide a model with additional information that allows for the association of a particular distribution with a set of contextual information. In this way, problems can be solved with a single contextualized model, rather than multiple models built for individual contexts.

This notion of context is a particularly useful observation when one considers the lack of labelled data inherent to novelty detection. In many novelty detection scenarios, for example seizure detection from EEG data, it is not practical to build individualized models for each context. It would therefore be useful to have have an effective way to build novelty detection models that can effectively utilize data from multiple contexts to build a single, accurate model. This motivates a semi-supervised approach where the detection algorithm is \textit{conditioned} on context, and therefore context-aware. This Context-Aware Novelty Detection autoEncoder (CANDE) therefore has two sources of information that lead it to become semi-supervised: the underlying assumption that all data in the training set is from the ``normal" class, along with a piece of information encoding the context of the normal data. In this paper we use these sources of information to build a novel contextually conditioned deep autoencoder model for context-aware novelty detection.

We first experimentally show on two datasets, including a large real world dataset, that shifts in context degrade detection ability necessitating individual models to be trained.  We then describe how conditioning deep autoencoders with easily obtained auxiliary labels can allow a single detector to be trained even on data displaying such contextual shifts. To incorporate this contextual information, we compare two methods of conditioning and propose the use of embeddings extracted from a context discriminator model which can obtain a more fine grained feature representations. With regard to these embeddings, we also analyse the role of embedding size on the performance of CANDE models. Our experiments reveal that similar, and sometimes even improved, performance to that achieved by multiple deep autoencoder models trained for different contexts can be achieved by a single CANDE model.

The remainder of the paper proceeds as follows. Section \ref{related_work} describes existing work related to deep novelty and anomaly detection in both static and streaming data as well as work on neural network conditioning. A detailed explanation of our proposed approach is provided in Section \ref{proposed_architecture}. We define the datasets used in our experiments, along with training procedures and model configurations used, in Section \ref{experimental_setup}. The results of our evaluation are presented in Section \ref{results}, along with a discussion of these results. Finally, Section \ref{conclusion} summarises our main findings and proposes directions for future work. 
\section{Related work}
\label{related_work}

This section describes related work in deep anomaly and novelty detection. We also describe work on contextual conditioning that motivates the methodology used in our proposed approach, along with work that addresses novelty or anomaly detection using deep architectures in the presence of shifts in distribution.

\subsection{Deep Anomaly Detection}
There is a significant amount of existing work in the area of anomaly and novelty detection in the context of deep learning. Here, it should be noted that we treat anomaly detection as a subcategory of novelty detection. Golan et al. \cite{golan2018deep} use a discriminative learning strategy for detecting anomalies in images. They first apply transformations to the images and then train a network to not only discriminate between normal and abnormal images, but also between each transformation. For inference, each transformation is applied to a query image and the novelty signal is based on the network's ability to recognise these transformations. Pidhorskyi et al. \cite{pidhorskyi2018generative} combined adversarial losses with reconstruction error in order to compute the likelihood of samples being generated from an inlier distribution. Gong et al. \cite{Gong2019memorizing} use a memory augmented network in order to improve detection by limiting the network to learning only  prototypical normal data. This avoids the undesirable scenario where the network generalises so well that it succeeds in also learning anomalies. This is achieved by learning a fixed number of sparse representations of the normal data via a content addressable memory, and retrieving these representations at test time via an attention mechanism. This way, only normal representations learned at training time can be retrieved, resulting in a poor reconstruction error for anomalous queries. Nguyen et al. \cite{nguyen2018anomaly}  addressed the issue of high likelihood being assigned to anomalous regions by encouraging a variational autoencoder to learn multi-model distributions using multiple hypothesis networks \cite{rupprecht2017learning} along with a discriminator network in order to prevent the network from assigning high likelihood to non-existent regions of the input. An important scenario in novelty detection that is not addressed well in the literature however, is the situation where data does not have a ``static" distribution for both normal and novel data. A very limited amount of research has been conducted on the subject of deep contextual anomaly or novelty detection. In terms of non domain specific models, \cite{shulman2019unsupervised} attempts to reconstruct contextual and behavioural attributes separately using variational autoencoders. There have also been a number of domain specific works on deep contextual models which will be detailed in the next section.

\subsection{Deep Anomaly Detection in Data Streams}
As our work relates to data exhibiting dynamic changes in distribution, it is pertinent to discuss deep learning approaches to novelty detection that specifically address streaming data. A continual learning approach was taken by Wiewel et al. \cite{wiewel2019continual} for scenarios where past data is unavailable when retraining models on new incoming data. They utilise generative replay \cite{shin2017continual} in order to augment their dataset, mitigating the effects of catastrophic forgetting when training on the most recent examples. 

Many of the works in this area are based on specific applications. For the application of anomaly detection in smart buildings, historical sensor data alongside contextual features were used to train autoencoders in \cite{araya2016collective}. In the field of acoustic modeling, a scene-dependent acoustic event detector was proposed by Komatsu et al. \cite{komatsu2019scene}. This used \textit{I-vectors}, a low-dimensional embedding which uses factor analysis of the difference between a Universal Background Model and a short audio segment-specific model, as an additional input to a WaveNet \cite{oord2016wavenet} model. WaveNet is an autoregressive model that also contains a conditioning mechanism. 
This work is perhaps the closest to addressing the problem addressed in this paper. Our proposed approach, however, is not domain-specific and differs in both the method of embedding and the conditioning mechanism used. We propose to use contextual information to augment a deep autoencoder model, and we suggest doing this in two ways: firstly by incorporating labels that denote contextual information, and secondly by using an embedding from a context discriminator model that gives more fine grained feature information for each context. We incorporate this information through the use of conditioning. The next section describes the relevant background on conditioning.

\subsection{Contextual Conditioning}
\textit{Conditioning} in deep learning  has provided an efficient mechanism with which to manipulate the output of deep networks such as generative adversarial networks \cite{radford2015unsupervised}, autoreggressive networks \cite{oord2016wavenet}, and  variational autoencoders \cite{sohn2015learning}. Conditioning works by using auxiliary inputs to apply transformations to the activations of an existing network. This mechanism has been applied over a wide range of domains from image  generation \cite{van2016conditional} and style transfer \cite{Dumoulin2017alearned}, to speech synthesis \cite{oord2016wavenet} and source separation \cite{slizovskaia2019end}. Though many works apply conditioning, the manner in which this conditioning is applied tends to vary, with both shifting and scaling of activations being common. Moreover, the layer at which these operations occur also varies throughout the literature. Methods include conditional batch normalisation \cite{de2017modulating} which learns the shifting and scaling parameters of batch normalization using conditioning information.  Feature-wise sigmoidal gating \cite{dumoulin2018feature-wise,oord2016wavenet} on the other hand, passes conditional information through one or more network layers and passes the resulting representations through a sigmoid function, leading to an output between 0 and 1. This is then used to conditionally 'gate' activations. An all-encompassing approach, termed \textit{Feature-wise Linear Modulation} (FiLM), was proposed by Perez et al. \cite{perez2018film}. FiLM uses both scaling and shifting operations while empirically showing that these operations, originally proposed in conditional batch normalisation \cite{de2017modulating}, could be effectively decoupled from batch normalisation operations, and applied at all layers in a network.

\section{Context-Aware Novelty Detection autoEncoder}
\label{proposed_architecture}
We propose a novel Context-Aware Novelty Detection autoEncoder (CANDE) that dynamically adapts the output of a standard deep novelty detection approach to contextual attributes. The advantage of this architecture is that it allows a single model to be trained for disparate contexts. The architecture is comprised of two components: a deep autoencoder and a contextual encoding function. These are described in detail in the following sections. 

\subsection{Deep Autoencoder}
Autoencoders, in their simplest form, learn meaningful representcations by using an encoding function to compress an input feature vector that is then decoded with a decoding function in order to ``reconstruct" the input. Deep autoencoders are created by simply increasing the number of layers in the encoder and decoder. Formally we define a fully connected autoencoder (AE) to have inputs $\pmb{x} \in \mathbb{R}^d$ and output ``reconstruction'' $\pmb{x}^\prime\in\mathbb{R}^d $, where $d$ refers to the input dimension. We define an encoder network $f: \mathbb{R}^d \to \mathbb{R}^r$, where $r$ is the dimension of the encoding $\pmb{z}$; and a decoder network $g: \mathbb{R}^r \to \mathbb{R}^d$. The encoder is therefore defined as 
\begin{equation}
    \pmb{z} = f(\pmb{x}, \theta^e)
\end{equation}
where $\theta^e$ defines the parameters of the encoder network. While the decoder is defined as 
\begin{equation}
    \pmb{x'} = g(\pmb{z}, \theta^d)    
\end{equation}
\noindent where $\theta^d$ defines the parameters of the decoder network. In the proposed model the encoder and decoder parameters, $\theta^e$ and  $\theta^d$, are untied \footnote{See \cite{vincent2010stacked} for more information about tied versus untied parameters in autoencoders}. The parameters of the autoencoder are trained by minimizing the reconstruction error between the input $\pmb{x}$ and the reconstruction $\pmb{x}'$.
 
For novelty detection, autoencoders are trained on normal data with the expectation that, at test time when query data is presented to the network, the reconstruction error will be higher for novel data than for normal data. This means that reconstruction error is used as a measure of novelty. 


\subsection{Conditioned Autoencoder}
To encourage an autoencoder to adapt to contextual information, and thereby modulate its output depending on a given context, the network is conditioned using an auxiliary context label. From a high level, the auxiliary context label denotes some partitioning of the dataset which indicates the context from which the data was derived. For instance, this could be a label indicating the day on which traffic volumes were recorded or the environment in which audio was recorded. Context is recorded in a training dataset through a set of auxiliary labels, $\langle C_i \rangle_{i \in I}$. Crucially, these labels do not contain any information about the nature of novelties but only serve as a more fine-grained representation of each normal example used to train the network. 

To condition the deep autoencoder in CANDE, we use the Feature-wise Linear Modulation (FiLM) \cite{perez2018film} conditioning strategy.\footnote{This layer-by-layer conditioning is also in contrast with a previous approach to autoencoder conditioning proposed by \cite{rudy2014generative} where simply the layer before the "bottleneck" is conditioned using a gating procedure for the purposes of sampling conditional distributions.} This approach avoids the difficult choice regarding the most appropriate layer to apply conditioning by simply conditioning all layers.  An affine transformation is applied the $k$th layer of the network, denoted by $\pmb{z}_k \in \mathbb{R}^q$, using shifting and scaling factors $\pmb{\gamma}_k$ and $\pmb{\beta}_k$. These factors are derived using the context vector $\pmb{h}_c \in \mathbb{R}^p$ as follows \cite{perez2018film}. 

\begin{equation}\label{gamma}
    \pmb{\gamma}_k = \pmb{h}_cW_{\gamma k} + \pmb{b}_{\gamma k}
\end{equation}

\begin{equation}\label{beta}
    \pmb{\beta}_k = \pmb{h}_cW_{\beta k} + \pmb{b}_{\beta k}
\end{equation}
\noindent where $W_{\gamma k}$, $W_{\beta k}$ are weight matrices corresponding to $\pmb{\gamma}_k$ and $\pmb{\beta}_k$, and $\pmb{b}_{\gamma k}$ and $\pmb{b}_{\beta k}$ represent their respective bias vectors.
The context vector $\pmb{h}_c$ can represent a one-hot-encoding of the context label or a more complex representation, such as an embedding. These transformations in equations \ref{gamma} and \ref{beta} are performed such that their output will be of dimension $q$, thus aligning with that of the layer to be conditioned, $\pmb{z}_k$. As $\pmb{\gamma}_k$ and  $\pmb{\beta}_k$ are now the same dimension as $\pmb{z}_k$ conditioning is then performed using an element-wise affine transformation \cite{perez2018film}.

\begin{equation}\label{z_prime}
    \pmb{z}_k' = \pmb{\gamma}_k\odot\pmb{z}_k + \pmb{\beta}_k 
\end{equation}

\begin{figure}[!tb]
  \centering
\includegraphics[width=\linewidth]{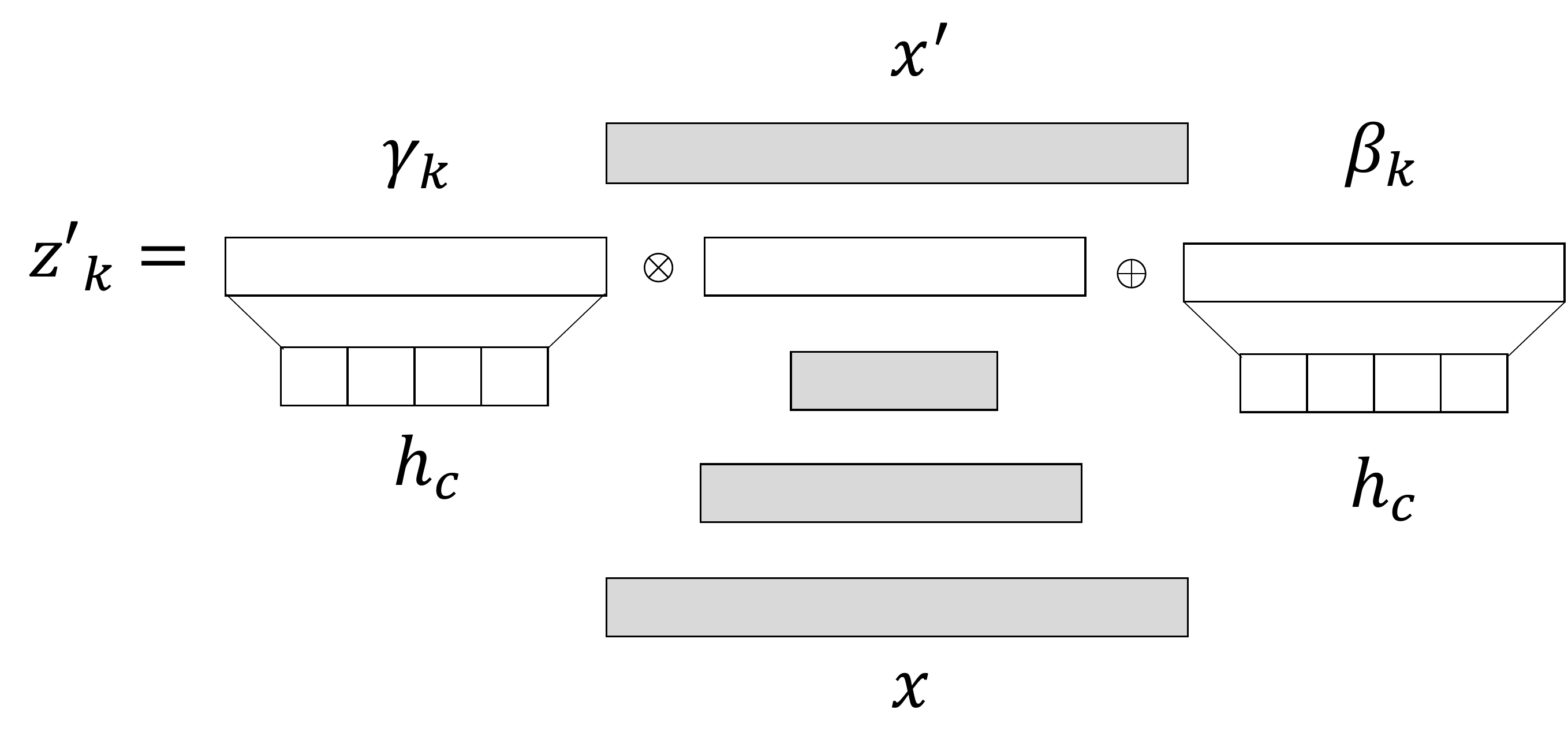}
  \caption{For a layer $k$, a label corresponding to the context of a particular example is transformed according to equations \ref{gamma} and \ref{beta}. The conditioned layer corresponds to the layer after being scaled by $\pmb{\gamma}_k$ and shifted by $\pmb{\beta}_k$. These operations occur at each layer in the network.}  \label{fig:conditioning_layer}
\end{figure}

\noindent After this transformation  the output of this layer, $\pmb{z}_k'$, is passed through a ReLU \cite{glorot2011deep} non-linearity. The conditioning process is illustrated in figure \ref{fig:conditioning_layer}.

\subsection{Contextual Encoder}

We explore two ways to encode the  contextual information into a conditioning vector, $\pmb{h}_c$. First, we explore the most obvious approach of using a one-hot-encoded vector of the context label as the conditioning vector, $\pmb{h}_c$. When a one hot encoded vector is used for conditioning, given a one-hot-encoded vector $\pmb{c}_i$ for the $i^{th}$ context,  $\pmb{h}^c = \pmb{c}_i$. 

Second, we explore the use of a conditioning vector derived from a discriminative model trained to discriminate between different contexts, a \emph{context discriminator model}.
We define a discriminative model, $d(\pmb{x})$, which predicts the context to which a training example, $\pmb{x}$, belongs. The predicted context, $\hat{c}$, is defined as

\begin{equation}
    \hat{c} = d(\pmb{x}, \theta^c)
\end{equation}

\noindent where $d(\pmb{x};\theta^c)$ is a fully connected neural network with $k$ layers, and $\theta^c$ are the network parameters which are comprised of a set of weight matrices $\theta^c=\{W^c_1, ..., W^c_k\}$, one for each layer of the network. 

In order to condition the deep autoencoder on the features learned by the discriminative model, the output of the penultimate layer in the context discriminator model, layer $k-1$, is used as the conditioning vector, $\pmb{h}_c$. The intuition behind the use of this layer in particular is that deep networks have been shown to learn features that are more and more linearly separable as the network increases in depth. This idea has been experimentally reinforced by \cite{alain2016understanding}.

\begin{figure}[!tb]
  \centering
\includegraphics[width=\linewidth]{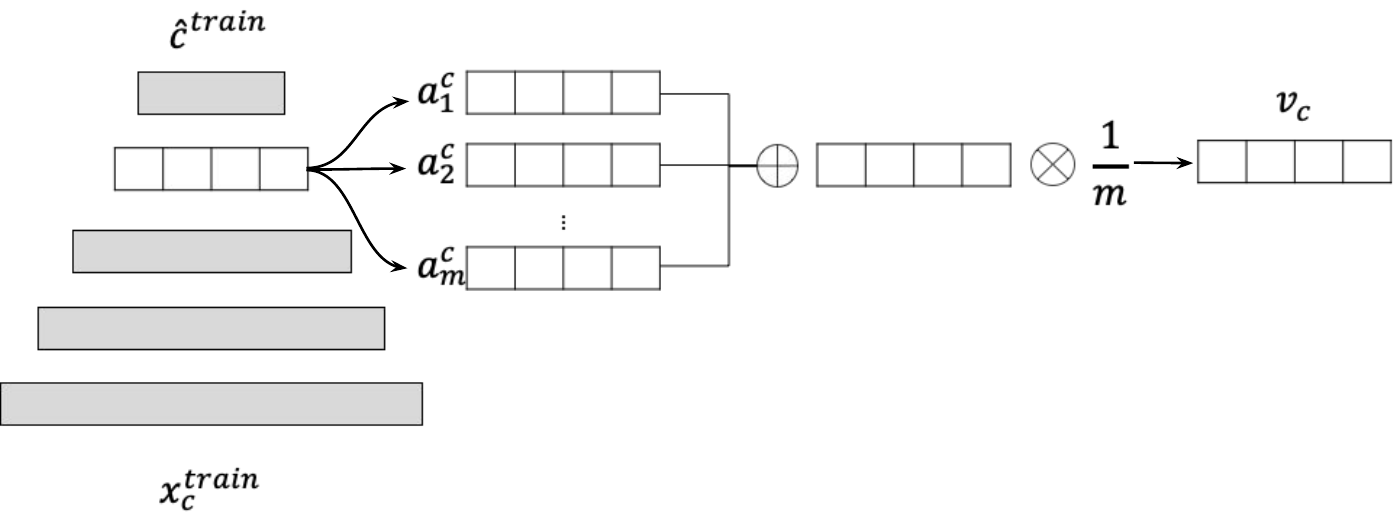}
  \caption{A conditioning vector generated by taking an average of the embeddings generated at the penultimate layer of the discriminator network for all examples in the training. }\label{fig:conditioning_vector}
\end{figure}

The most obvious way to extract the conditioning vector from the discriminator network would be to feed training examples into the network and use the activations of the penultimate layer directly to condition the autoencoder. However, at testing time, it will not be known whether query examples will be of the normal or novel class (distinguishing between these classes is the point of the autoencoder!). If conditioning vectors based on embeddings generated in the discriminator network for novel examples are used, it is likely that they would not be representative of the context in which the novel example resides. 

An alternative approach that we take is to create an embedding for each context that utilises the activations obtained during the training of the discriminative model. In this case, the activations for each context are collected during training and the mean vectors of these activations are used as an embedding for each context. More formally, the activation of the penultimate layer $\pmb{a}_i \in \mathbb{R}^p$ for a given input $\pmb{x}_i$ is defined as 
\begin{equation}
    \pmb{a}_i =  d(\pmb{x}_i;\{W_1^c,...W_{l-1}^c\})
\end{equation}

We can then construct an $n\times p$ matrix, $H$, where $n$ is the number of training examples in the training set and $p$ is the length of the penultimate layer vector $\pmb{a}_i$. For each example $\pmb{x}_i$ in the training set, there is a corresponding context $c_i$. If we take all rows where the corresponding context is equal to a context $c$, this leads to a submatrix $H_c$. The embedding vector $\pmb{h}_c$ consists of the mean vector of the columns of $H_c$. This process is illustrsated in figure  \ref{fig:conditioning_vector}. The modified conditioned autoencoder model is therefore described as:

\begin{equation}\label{ae_h_encoder}
    \pmb{z}' = f(\pmb{x}, \pmb{h}_c; \theta^e)
\end{equation}
\begin{equation}\label{ae_h_decoder}
    \pmb{x}' = g(\pmb{z}', \pmb{h}_c; \theta^d)    
\end{equation}

This approach to generating conditioning vectors is similar to \textit{d-vectors} where the mean of activations of the last layer of a network is used for a number of speakers in order to build a representation of each speaker for the task of speech verification \cite{variani2014deep}. This strategy is also similar to a method proposed in \textit{prototypical networks} in the field of few-shot learning \cite{snell2017prototypical}, where a vector representing the mean of the embedded support set examples for each of a set of classes is used as a \textit{prototype} representation. This also bares a resemblance to a Universal Background Model (UBM) which is used as conditioning criteria by \cite{komatsu2019scene} on a WaveNet style anomaly detector. Though, in contrast, our representation is not domain specific and can, in theory, be used with any data type.

\section{Experiments}\label{experimental_setup}
We have designed a set of evaluation experiments to evaluate the performance of CANDE, and to measure the effect of using conditioning on novelty detection models. We compare CANDE with different conditioning criteria and use various unconditioned models as benchmarks with which to evaluate the effectiveness of conditioning. All hyperparameters information and dataset details can be found in sections 1 and 2 of the supplementary material. \footnote{The code required to reproduce these experiments is available at \url{https://github.com/EllenRushe/CANDE.git}}. 

\subsection{Data}
We use two datasets in these experiments: a novelty detection dataset based on MNIST  and the MIMII dataset which is a large dataset that captures a real-world novelty detection scenario using audio recorded from industrial machines. 

\subsubsection{MNIST}
We first replicate contextual novelties artificially using the MNIST dataset \cite{lecun1998mnist}. This is done in order to mimic the phenomenon of shifting contexts. The MNIST dataset  is partitioned into different ``contexts". A context denotes a set of classes from the dataset. This set of classes is considered ``normal" for that context.  Let $ \langle C_i \rangle_{i \in I}$ denote a set of different contexts in a dataset, where 
$\forall{i,j}\in I : i\neq j \Rightarrow C_i \cap C_j = \O$.  The discriminator model discriminates between the set of contexts $C$. For example, in the case where $|C| = 3$, $c_n$ is the context label and $y_n$ denotes the digit class label of example $n$. 
\begin{equation}
c_n = 
    \begin{cases}
      100 & y_n \in C_1\\    
      010 & y_n \in C_2  \\
      001 & y_n \in C_3
    \end{cases}
\end{equation}

\begin{figure}[h]\label{mnist_picture}
  \centering
\includegraphics[width=\linewidth]{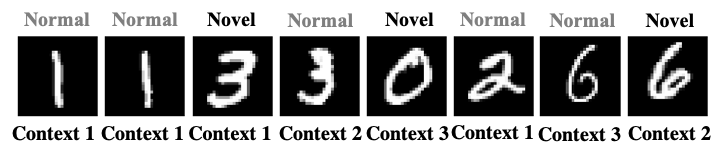}
  \caption{Example of MNIST dataset configuration}
\end{figure}

For MNIST, we partition the data into three distinct contexts where the labels in each context do not overlap. These are: $C_1=\{0,1,2\}$, $C_2=\{3,4,5\}$ and $C_3=\{6,7,8\}$. 
To simulate contextual novelties, we then relabel data as being `normal' or `novel' and create scenarios where in one context, images of a specific digit are `novel', whereas in another context images of the same digit are considered `normal'. Figure \ref{mnist_picture} illustrates this.
When the different contexts are combined, there will be images of the same digit with both `novel' and `normal' labels in the dataset, as they belong to different contexts. This is a somewhat unrealisticly challenging situation given the extreme overlap between the normal and novel distributions but this is done on purpose to clearly demonstrate the effects of conditioning in a controlled manner. Conditioning should allow us to take the context into account therefore discriminating these seemingly similar types of data. Without conditioning, we expect that a traditional semi-supervised model would be unable to detect that these examples differ in any way.

\subsubsection{MIMII Dataset}
As an example of a real-world novelty detection problem, we use the recently proposed MIMII public dataset \cite{purohit2019mimii} which consists of industrial machine sounds recorded from a number of different machine types and models. The task requires novel audio to be detected in 10-second audio segments, where a novel audio segment might indicate a machine about to fail. The original dataset consists of eight channels of 16-bit audio sampled at 16 KHz. Background environmental noise was also mixed into the target machine sounds in order to make conditions more realistic. This background noise was added at different signal-to-noise (SNR) ratios, namely 6dB, 0dB and -6dB, leading to three levels of difficulty. As in the original benchmarks in \cite{purohit2019mimii}, in our experiments we use a representation of audio segments that consists of five frames of 64 log-mel spectrogram filters derived from the first channel of audio with a frame size of 1,024 and a hop-length of 512. The same test data configuration was used as in \cite{purohit2019mimii} with all anomalous sounds being used in the test set  along with an equal number of normal segments. This left the remainder of the data to be used for training and validation, of which 90\% was used for training and 10\%  for validation. In the public MIMII dataset four different types of machine were recorded: `valve', `pump', `fan' and `slide rail'. Recordings from four different models of each machine are included. Therefore there are 16 individual machine models, each of which we consider to define a context.

\subsection{Models}
We compare a number of different baselines and variations of CANDE:

\noindent\textbf{Individual Models:} For each context, an individual autoencoder model was trained in order to ascertain the best performance possible given a model trained on a particular context, without intrusion of other normal contexts. 

\noindent\textbf{Combined Models:} Combined models take all data from all contexts and combine them in order to train a single novelty detection model. In the case of MNIST this means that, in the test set, there will be instances where two examples that are drawn from the same distribution (for instance two "0" digits) will have opposite labels. This represents an extreme overlap between normal and abnormal distributions. In the case of the MIMII data, this effect is more subtle, as the combined dataset simply includes examples from all machine models. There are three types of combined models.
\begin{enumerate}[label=(\roman*)]
   \item \textbf{Unconditioned Model} We use unconditioned combined models as a benchmark against which to compare CANDE models.
  \item \textbf{CANDE One-hot-encoded:}
For FiLM conditioning, the ID of each context was first used as a context vector with which to condition models.
  \item \textbf{CANDE Contextual Embedding:} Contextual embeddings were derived from the activations of the context discriminator  model, one for each context, as described in Section \ref{proposed_architecture}.  Though the effect of embedding size on the proposed architecture is not fully understood, it is important to note that contextual embedding size  has been shown to have an effect on performance of conditioned models in applications of natural language processing \cite{melamud-etal-2016-role}. To investigate whether the embedding size has an effect for context embeddings, the embedding size was varied by training a number of different context discriminator models with different capacities, all with different penultimate sized layers. The dimensions evaluated are 32, 64, 128 and 256.  
\end{enumerate}
For all autoencoder models, mean squared error (MSE) is used as the cost function during training. The stopping criteria for training was determined by selecting the epoch with lowest MSE on the validation set. The depth, number of hidden layers, number of nodes, batch-size and learning rate were kept constant over all the semi-supervised networks in order to effectively evaluate the performance of CANDE models in comparison to their unconditioned counterparts. For each model, training, validation, and testing were performed a number of times for different random weight initialisations: ten times for MNIST and three times for the MIMII dataset (which is  significantly larger). 

\subsection{Evaluation}
In order to calculate a novelty score, the MSE is calculated between a given test example, $\pmb{x}_i$, and its reconstruction, $\pmb{\hat{x}}_i$. We base our evaluation on the Area Under the the Receiver Operator Characteristic (ROC) Curve (AUC). This measures the ability of the models to generate accurate novelty scores without requiring a novelty classification threshold to be set. 
For MNIST, the AUC was simply calculated for each individual context. For the MIMII dataset, evaluation was done in line with previous baselines using this dataset \cite{purohit2019mimii}. For every machine type and model, there were a number of audio files in the test set, half with label  `normal' and the other half with label `abnormal'. For each audio file of roughly 10 seconds, the MSE was calculated for each audio segment within that file, leading to a set of MSE values for each file. The mean of this set of MSE values was then used as the novelty score for this file. This was repeated for all files. The AUC was then calculated over all files for each context. 

\section{Results \& Discussion}
\label{results}
In this section we discuss the results of our experiments on the MNIST and MIMII datasets as well as an investigation of the impact of embedding size on the effectiveness of the CANDE models. 

\subsection{MNIST Dataset}
Table \ref{mnist_results_table} details the results for the MNIST-based dataset. The performance of each approach is ranked and average ranks across the three contexts are calculated to summarise performance. The separate models are omitted from this ranking as they will always achieve the top rank. Although in all cases but individual models only one model is trained, performance is measured separately for each context to give a more detailed view of the differences in model performance. The separate autoencoder models trained for each context perform very well however there is a huge degradation in performance in the combined model trained without conditioning. This is not surprising given the degree of overlap between the novel and normal classes in the three different contexts. 
Comparing the results of the CANDE conditioned combined models to the unconditioned combined model shows the clear improvement provided by contextual conditioning, and demonstrates that performance similar to that achieved with three separate models is possible using a single detector with conditioning. This is the case even when a simple one-hot encoded label is used as the conditioning vector. In fact, the model using this simple conditioning vector is the best performing model in this experiment. The performance of the CANDE models that use the embedded conditioning vector, although much better than that of the unconditioned model, is not better than the CANDE models using the simpler conditioning vector. It is likely that the reason that the simple conditioning vector works so well is that the degree of contradiction between the three contexts is so severe. It is, however, salient that all CANDE models exhibit performance quite close to that of the independently trained individual models.

\begin{table*}[h] \label{mnist_results_table}
\caption{This table shows the AUC and average rank om MNIST dataset.}
\centering \resizebox{0.8\linewidth}{!}{%

\begin{tabular}{l l r c c c c c c }
\hline  
\begin{tabular}[l]{@{}l@{}}Normal\\ digits\end{tabular} & \begin{tabular}[l]{@{}l@{}}Novel\\ digit\end{tabular} & \begin{tabular}[r]{@{}l@{}}AE individual\\models\end{tabular} & \begin{tabular}[r]{@{}r@{}}AE \\ no cond.\end{tabular} & \begin{tabular}[r]{@{}r@{}}CANDE \\ one hot\end{tabular} & \begin{tabular}[r]{@{}r@{}}CANDE \\ 32 embed\end{tabular} & \begin{tabular}[r]{@{}r@{}}CANDE\\ 64 embed\end{tabular} & \begin{tabular}[r]{@{}r@{}}CANDE \\ 128 embed\end{tabular} & \begin{tabular}[r]{@{}r@{}}CANDE \\ 256 embed\end{tabular} \\ \hline
0, 1, 2     & ~3      & 0.945     & 0.609     & 0.921  &  0.860   & 0.865   & 0.869      & 0.926  \\
3, 4, 5     & ~6      & 0.944     & 0.491     & 0.916  &  0.788   & 0.802   & 0.825      & 0.830  \\
6, 7, 8    & ~0       & 0.935     & 0.518     & 0.893  &  0.804   & 0.787   & 0.816      & 0.874                                                        \\ \hline
\multicolumn{2}{c}{Average Rank} && 6.000     & 1.333  &  4.667   & 4.333   & 3.000      & 1.667                                                        \\ \hline
\end{tabular}
}
\end{table*}

\subsection{MIMII Dataset}
The MIMII dataset provides a more realistic view of novelty detection in real world scenarios with data arising from a number of sources with varying degrees of degradation of performance when using combined models. Table \ref{mimii_all_results} shows the performance of each modelling approach. We again report results broken down by machine model. This allows us to get a more complete picture of model performance. To summarize results, each model has been ranked from 1 to 6, from best to worst, with the average rank across all contexts being used as a summary of the performance of each model. For brevity sake, able table \ref{mimii_all_results} reports the average performance across three signal-to-noise ratios, however ranks are computed across the full set of results which can be found in section 3 of the supplementary material. 

The need for conditioning is clearly illustrated by the degradation between the performance of the separate models and the performance of the individual models without conditioning. In all but a few cases, CANDE models outperform the unconditioned models. Unlike in the case of MNIST, the real-world, complex data seems to benefit from a more fine-grained representation of each context with models conditioned with embeddings being the top two performing models. Only in the case where the embedding size was 256, was the embedding conditioned model lower than the model conditioned on one-hot labels. For all context-aware models, the rank is lower than the unconditioned models, showing a clear performance increase gained from context awareness.

\begin{table*}[h]\label{mimii_all_results}
\caption{This table shows the AUC for MIMII dataset for all models. AUCs are calculated for each machine type and ID at three different signal-to-noise ratios (SNR).}
\centering \resizebox{0.8\linewidth}{!}{%
\begin{tabular}{llrcccccc}
\hline
Model                   & ID & \begin{tabular}[c]{@{}r@{}}AE individual\\models\end{tabular} & \begin{tabular}[c]{@{}c@{}}AE \\ no cond.\end{tabular} & \begin{tabular}[c]{@{}c@{}}CANDE \\ one hot\end{tabular} & \begin{tabular}[c]{@{}c@{}}CANDE \\ 32 embed\end{tabular} & \begin{tabular}[c]{@{}c@{}}CANDE \\ 64 embed\end{tabular} & \begin{tabular}[c]{@{}c@{}}CANDE \\ 128 embed\end{tabular} & \begin{tabular}[c]{@{}c@{}}CANDE\\  256 embed\end{tabular} \\ \hline
\multirow{4}{*}{fan}    & 00 & 0.663                                                              & 0.614                                                  & 0.647                                                      & 0.653                                                       & 0.649                                                       & 0.659                                                        & 0.625                                                        \\
                        & 02 & 0.850                                                              & 0.747                                                  & 0.845                                                      & 0.847                                                       & 0.850                                                       & 0.854                                                        & 0.784                                                        \\
                        & 04 & 0.748                                                              & 0.714                                                  & 0.764                                                      & 0.774                                                       & 0.776                                                       & 0.758                                                        & 0.674                                                        \\
                        & 06 & 0.930                                                              & 0.773                                                  & 0.904                                                      & 0.957                                                       & 0.938                                                       & 0.959                                                        & 0.879                                                        \\ \hline
\multirow{4}{*}{pump}   & 00 & 0.609                                                              & 0.466                                                  & 0.626                                                      & 0.572                                                       & 0.600                                                       & 0.611                                                        & 0.691                                                        \\
                        & 02 & 0.519                                                              & 0.426                                                  & 0.430                                                      & 0.432                                                       & 0.409                                                       & 0.451                                                        & 0.474                                                        \\
                        & 04 & 0.950                                                              & 0.737                                                  & 0.938                                                      & 0.944                                                       & 0.951                                                       & 0.952                                                        & 0.908                                                        \\
                        & 06 & 0.805                                                              & 0.647                                                  & 0.775                                                      & 0.798                                                       & 0.792                                                       & 0.763                                                        & 0.761                                                        \\ \hline
\multirow{4}{*}{slider} & 00 & 0.973                                                              & 0.968                                                  & 0.964                                                      & 0.968                                                       & 0.964                                                       & 0.967                                                        & 0.956                                                        \\
                        & 02 & 0.860                                                              & 0.796                                                  & 0.847                                                      & 0.861                                                       & 0.865                                                       & 0.832                                                        & 0.832                                                        \\
                        & 04 & 0.765                                                              & 0.755                                                  & 0.840                                                      & 0.848                                                       & 0.863                                                       & 0.801                                                        & 0.764                                                        \\
                        & 06 & 0.625                                                              & 0.668                                                  & 0.636                                                      & 0.647                                                       & 0.653                                                       & 0.600                                                        & 0.578                                                        \\ \hline
\multirow{4}{*}{valve}  & 00 & 0.540                                                              & 0.362                                                  & 0.507                                                      & 0.491                                                       & 0.479                                                       & 0.475                                                        & 0.404                                                        \\
                        & 02 & 0.619                                                              & 0.647                                                  & 0.635                                                      & 0.660                                                       & 0.653                                                       & 0.618                                                        & 0.623                                                        \\
                        & 04 & 0.625                                                              & 0.504                                                  & 0.594                                                      & 0.573                                                       & 0.582                                                       & 0.553                                                        & 0.546                                                        \\
                        & 06 & 0.651                                                              & 0.565                                                  & 0.590                                                      & 0.596                                                       & 0.607                                                       & 0.616                                                        & 0.570                                                        \\ \hline
\multicolumn{3}{l}{Average rank}                                                                  & 5.083                                                  & 3.271                                                      & 2.688                                                       & 2.646                                                       & 3.083                                                        & 4.229                                                        \\ \hline
\end{tabular}

}
\end{table*}

\subsection{Exploring Embedding Size}
\label{discussion}
Our results show that CANDE provides a clear advantage over models lacking this auxiliary information. It is also clear, however, that the size of the contextual embedding used has an impact on model performance. To explore the possibility of determining an optimal embedding size based on the performance of the context discriminator, we evaluated the discriminator with different embedding sizes using a validation set. Note we evaluate on the the training epoch that provides the best validation set accuracy as this is the epoch from which the embeddings were generated. 

From Table \ref{mimii_val_acc}, we can see that the best validation accuracy is obtained using the model with penultimate layer size 64, while the worst accuracy results from that with 256. This correlates with the rankings of performance in the novelty detection problem, suggesting that there may be a connection between the validation accuracy obtained on the network from which the embedding is obtained and the effectiveness of this vector in representing a particular context.

\begin{table}[h]\label{mimii_val_acc}
\caption{Accuracy on validation accuracy for all discriminative models with penultimate layers of various sizes.}
\centering
\begin{tabular}{cc}
\hline
Layer size & \multicolumn{1}{c|}{\% Validation Accuracy} \\ \hline
32         & 82.89                                       \\
64         & \textbf{83.05}                              \\
128        & 83.02                                       \\
256        & 82.45                                       \\ \hline
\end{tabular}

\end{table}

It is not exactly clear if this is the case, however the performance of the networks with embedding size 32 and 128 does not predict validation accuracy ranking, though the differences are minimal. It is also not clear whether the increase in performance arises from the size of the embedding, or the accuracy of the discriminative network itself. For instance the higher dimensional discriminative network only uses early stopping for regularisation and may simply be overfitting the training data due to the increased capacity of the network. Deeper analysis on this issue is left for future enquiry.

\section{Conclusion \& Future Work}
\label{conclusion}
In this paper we have proposed a contextually conditioned deep autoencoder for context-aware novelty detection. We utilise contextual encodings both in the form of auxiliary contextual labels and in the form of embeddings in order to condition a model to modulate its outputs based on context. We have shown that context-aware architectures clearly outperform their unconditioned counterparts in nearly all cases, especially where there is a high degree over overlap between normal and novel labels. We have demonstrated that utilising embeddings derived from discriminative models is effective for conditioning such models, specifically in the case of the complex real-world problem of acoustic anomaly detection. We also analyse the effect of embedding size on performance. The results demonstrate that this architecture can recover much of the performance lost by training a single model on all data combined and in some cases can even out-perform individually trained models. The main advantage of using CANDE is also well illustrated in the real-world scenario where very similar performance to that achieved using 16 separate models can be achieved with a single CANDE model.

 For future work, we plan to modify the current model to address more gradual shifts in context, i.e. concept drift, and investigate how entirely new contexts can be incorporated into the system. We will also explore multiple different  ways in which conditioning vectors can be generated and incorporated into the deep autoencoders.

\bibliographystyle{siamplain.bst}
\bibliography{bibliography.bib}

\end{document}